\documentclass{article}

\usepackage{microtype}
\usepackage{graphicx}
\usepackage{subfigure}
\usepackage{booktabs} 

\usepackage{hyperref}

\usepackage{multirow}
\usepackage{subfigure}
\usepackage{bm}
\usepackage{balance} 
\usepackage{amssymb}
\usepackage{wrapfig}
\usepackage{amsmath}
\usepackage{enumitem}
\usepackage{threeparttable}
\usepackage{xcolor}
\usepackage{adjustbox}

\usepackage[accepted]{icml2021}

\begin{document}

\twocolumn[
\icmltitle{Learning to Rehearse in Long Sequence Memorization}



\icmlsetsymbol{equal}{*}

\begin{icmlauthorlist}
\icmlauthor{Zhu Zhang}{equal,zju,ali}
\icmlauthor{Chang Zhou}{equal,ali}
\icmlauthor{Jianxin Ma}{ali}
\icmlauthor{Zhijie Lin}{zju}
\icmlauthor{Jingren Zhou}{ali}
\icmlauthor{Hongxia Yang}{ali}
\icmlauthor{Zhou Zhao}{zju}

\end{icmlauthorlist}

\icmlaffiliation{ali}{DAMO Academy, Alibaba Group, China}
\icmlaffiliation{zju}{Zhejiang University, China}

\icmlcorrespondingauthor{Zhou Zhao}{zhaozhou@zju.edu.cn}

\icmlkeywords{Machine Learning, Long Sequence Memorization, Memory-Based Reasoning, Rehearsal}

\vskip 0.3in

]



\printAffiliationsAndNotice{\icmlEqualContribution} 

\begin{abstract}
Existing reasoning tasks often have an important assumption that the input contents can be always accessed while reasoning, requiring unlimited storage resources and suffering from severe time delay on long sequences. To achieve efficient reasoning on long sequences with limited storage resources, memory augmented neural networks introduce a human-like write-read memory to compress and memorize the long input sequence in one pass, trying to answer subsequent queries only based on the memory. But they have two serious drawbacks: 1) they continually update the memory from current information and inevitably forget the early contents; 2) they do not distinguish what information is important and treat all contents equally. In this paper, we propose the Rehearsal Memory (RM) to enhance long-sequence memorization by self-supervised rehearsal with a history sampler. To alleviate the gradual forgetting of early information, we design self-supervised rehearsal training with recollection and familiarity tasks. Further, we design a history sampler to select informative fragments for rehearsal training, making the memory focus on the crucial information. We evaluate the performance of our rehearsal memory by the synthetic bAbI task and several downstream tasks, including text/video question answering and recommendation on long sequences.
\end{abstract}

\section{Introduction}
In recent years, the tremendous progress of neural networks has enabled machines to perform reasoning given the input contents $X$ and a query $Q$, e.g., infer the answer of given questions from the text/video stream in text/video question answering~\citep{seo2016bidirectional,jin2019multi,le2020hierarchical}, or predict whether a user will click the given item based on the user behavior sequence in recommender systems~\citep{ren2019lifelong,pi2019practice,zhang2021cause}.
Studies that achieve top performances at such reasoning tasks usually have an important assumption that the raw input contents $X$ can be always accessed while answering the query $Q$.
In this setting, the complex interaction between $X$ and $Q$ can be designed to extract query-relevant information from $X$ with little loss, such as co-attention interaction~\citep{xiong2016dynamic,jin2019video}. Though these methods~\citep{seo2016bidirectional,le2020hierarchical} can effectively handle these reasoning tasks, they require unlimited storage resources to hold the original input $X$. Further, they have to encode the whole input contents and develop the elaborate interaction from scratch, which are time-consuming. This is not acceptable for online services that require instant response such as recommender systems, as the input sequence becomes extremely long~\citep{ren2019lifelong}.

To achieve efficient reasoning on long sequences with limited storage resources, memory augmented neural networks~(MANNs)~\citep{graves2014neural,graves2016hybrid} introduce a write-read memory $M$ with fixed-size capacity~(size much smaller than $|X|$) to compress and remember the input contents $X$. In the inference phase, they can capture query-relevant clues directly from the memory $M$, i.e., the raw input $X$ is \emph{not} needed at the time of answering $Q$.
This procedure is very similar to the daily situation of our human beings, i.e., we may not know the tasks $Q$ that we will answer in the future when we are experiencing current events, but we have the instincts to continually memorize our experiences within the limited memory capacity, from which we can rapidly recall and draw upon past events to guide our behaviors given the present tasks~\citep{moscovitch2016episodic,baddeley1992working}. 
Such human-like memory-based methods bring three benefits for long-sequence reasoning: 1) storage efficiency: we only need to maintain the limited memory $M$ rather than $X$; 2) reasoning efficiency: inference over $M$ and $Q$ is more lightweight than inference over $X$ and $Q$ from scratch ; 3) high reusability: the maintained memory $M$ can be reused for any query $Q$.

However, existing MANNs have two serious drawbacks for memory-based long-sequence reasoning. First, these approaches ignore the long-term memorization ability of the memory.
They learn how to maintain the memory $M$ only by back-propagated losses to the final answer and do not design any specific training target for long-term memorization, which inevitably lead to the gradual forgetting of early contents~\citep{le2019learning}. That is, when dealing with the long input sequences, these approaches may fail to answer the query relevant to early contents due to the lack of long-term memorization.
Second, determining what to remember in the memory with limited capacity is crucial to retain sufficient clues for subsequent $Q$. This is especially challenging since the information compression procedure in $M$ is totally not aware of $Q$.
But existing MANNs do not distinguish what information is important and treat all contents equally. Thus, due to lack of information discrimination, these approaches may store too much meaningless information but lose vital evidence for subsequent reasoning.

In this paper, we propose the Rehearsal Memory (RM) to enhance long-sequence memorization by self-supervised rehearsal with a history sampler. 
To overcome gradual forgetting of early information and increase the generalization ability of the memorization technique, we develop two extra self-supervised rehearsal tasks to recall the recorded history contents from the memory. The two tasks are inspired by the observation that human beings can recall details nearby some specific events and distinguish whether a series of events happened in the history, 
which respectively correspond to two different memory processes revealed in cognitive, neuropsychological, and neuroimaging studies, i.e., \emph{recollection} and \emph{familiarity}~\citep{yonelinas2002nature,moscovitch2016episodic}.
Concretely, the recollection task aims to predict the masked items in history fragments $H$, which are sampled from the original input stream and parts of items are masked as the prediction target. 
This task tries to endow the memory with the \emph{recollection} ability that enables one to relive past episodes.
And the familiarity task tries to distinguish whether a historical fragment $H$ ever appears in the input stream, where we directly sample positive fragments from the input stream and replace parts of the items in positive ones as negative fragments. 
This task resembles the \emph{familiarity} process that recognizes experienced events or stimulus as familiar.

To make the rehearsal memory have the ability of remembering the crucial information, we further train an independent history sampler to select informative fragments $H$ for self-supervised rehearsal training. Similar to the teacher-student architecture in knowledge distillation~\citep{hinton2015distilling}, we expect the history sampler (i.e. the teacher) to capture the characteristic of important fragments in the current environment and guide the rehearsal memory (i.e. the student) to remember task-relevant clues. Concretely, we independently train a conventional reasoning model that can access raw contents $X$ while answering the query $Q$ as the history sampler. The model contains the attention interaction between history fragments $H$ and the query $Q$, where the attention weight can be regarded as the importance of each fragment. After training, the history sampler can select the vital fragments based on the attention weights for self-supervised rehearsal training. This is similar to the procedure where human beings learn to memorize meaningful experiences, i.e., we have gone through a lot of tasks to slowly understand which information is likely to be used in future tasks and pay more attention to them during memorization~\citep{moscovitch2016episodic}.

In conclusion, we propose the self-supervised memory rehearsal to enhance the long-sequence memorization for subsequent reasoning. We design the self-supervised recollection and familiarity tasks to solve \emph{how to rehearse}, which can alleviate the gradual forgetting of early information.  Further, we adopt a history sampler to decide \emph{what to rehearse}, which guides the memory to remember critical information.
We illustrate the ability of our rehearsal memory via the synthetic bAbI task and several downstream tasks, including text/video question answering and recommendation on long sequences.

\section{Related Works}
Memory augmented neural networks (MANNs) introduce the external memory to store and access the past contents by differentiable write-read operators. Neural Turing Machine (NTM)~\citep{graves2014neural} and  Differentiable Neural Computer (DNC)~\citep{graves2016hybrid} are the typical MANNs for human-like memorization and reasoning, whose inferences rely only on the memory with limited capacity rather than starting from the original input. 
In this line of research, \citet{rae2016scaling} adopt the sparse memory accessing to reduce computational cost. \citet{csordas2019improving} introduce the key/value separation problem of content-based addressing and adopt a mask for memory operations as a solution. 
\citet{le2019neural} manipulate both data and programs stored in memory to perform universal computations.
And \citet{santoro2018relational,le2020self} consider the complex relational reasoning with the information they remember.

However, these works exploit MANNs mainly to help capture complex dependencies in dealing with input sequences, but do not explore the potential of MANNs in the field of memory-based long-sequence reasoning. 
They learn how to maintain the memory only by back-propagated losses to the final answer but do not design specific training target for long-term memorization, inevitably incurring gradual forgetting of early contents during memorizing long sequences~\citep{le2019learning}.
Recently, there are a few works trying to alleviate this problem. \citet{le2019learning} propose to measure ``remember'' ability by the final gradient on the early input, and adopt a uniform writing operation on the memory to balance between maximizing memorization and forgetting.
\citet{munkhdalai2019metalearned} design the meta-learned neural memory instead of the conventional array-structured memory and memorize the current and past information by reconstructing the written values via the memory function.
Besides, Compressive Transformer~\citep{rae2019compressive} maps the past memory to a smaller compressed memory for long-range sequence learning, where the compressed memory preserves much original information by a high compression rate. But considering the compressed memory is implemented by a FIFO queue, it will completely forget the contents beyond a fixed range.

Our approach is different and parallel to these techniques, we try to enhance long-sequence memorization by self-supervised memory rehearsal, i.e., recall the recorded history contents from the memory to overcome gradual forgetting of early information. We design the recollection task to enable the memory to relive past episodes and adopt the familiarity task to make the memory recognize experienced events.
A recent work~\citep{park2020distributed} also introduces a self-supervised memory loss to ensure how well the current input is written to the memory, but it only focuses on remembering the current information and ignoring the long-term memorization.
Further, compared to previous techniques that have no assumptions on what behavior will be remembered the most, we propose a history sampler to distinguish the characteristic of important fragments in the current environment and guide the memory rehearsal to remember task-relevant clues.

\section{Rehearsal Memory}

\begin{figure*}[th]
    \centering
    \includegraphics[width=1\textwidth]{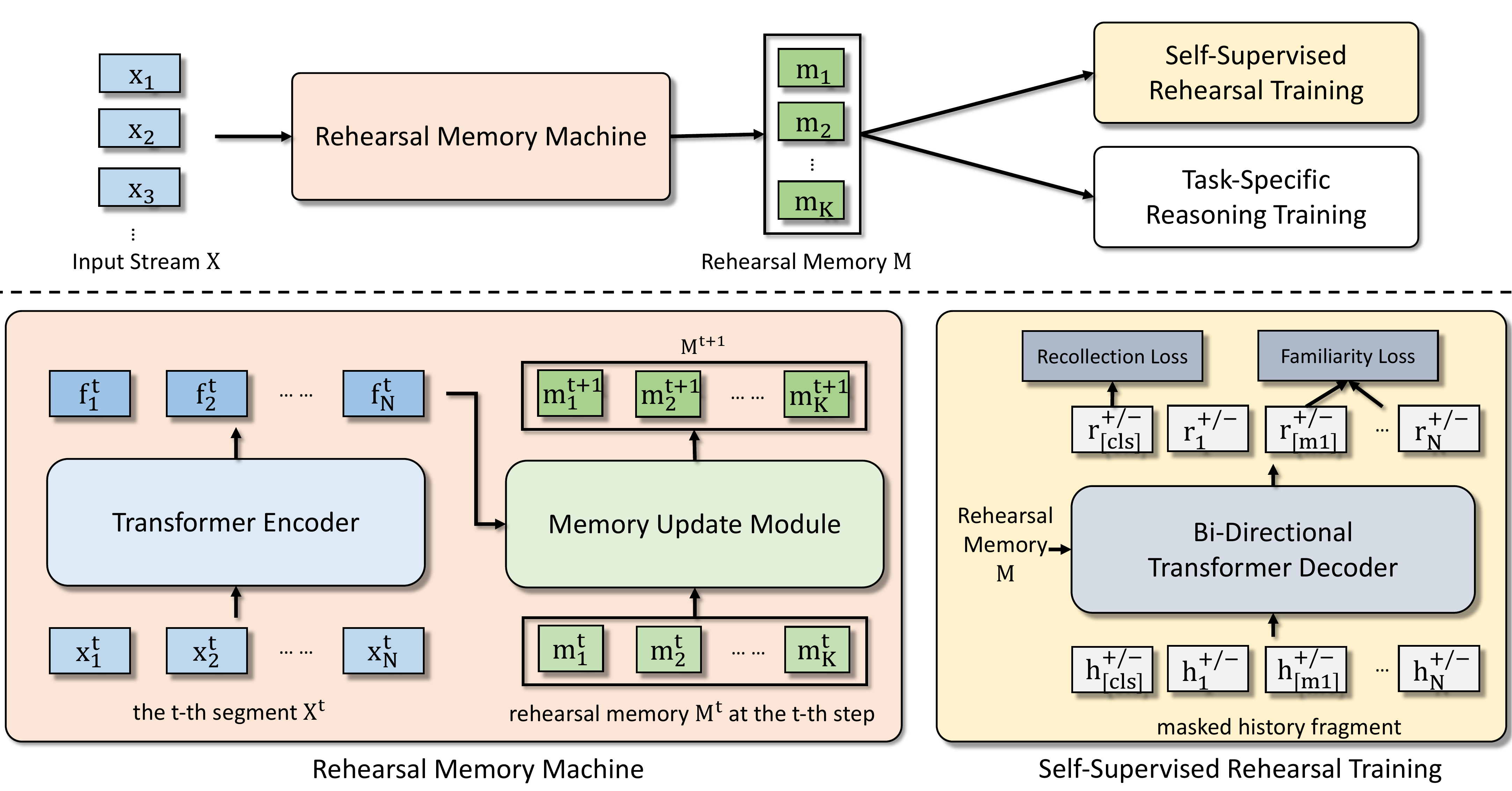} 
    \caption{ The Framework of Rehearsal Memory and Self-Supervised
Rehearsal Training.}
    \label{fig:framework}
\end{figure*}

\subsection{Problem Formulation}
Given the input stream ${\bf X} = \{{\bf x}_1, {\bf x}_2, \cdots\}$ and a query ${\bf Q}$, the directly reasoning methods~\citep{seo2016bidirectional,le2020hierarchical} learn the model ${\mathcal T}({\bf X},{\bf Q})$ to predict the answer ${\bf A}$. These is an important assumption that the input stream ${\bf X}$ can be always accessed while reasoning. And complex interaction between ${\bf X}$ and ${\bf Q}$ can be designed to extract query-relevant information in ${\mathcal T}({\bf X},{\bf Q})$. Obviously, these methods have to store the original input ${\bf X}$ and infer the answer ${\bf A}$ from scratch when the query ${\bf Q}$ is known.
In this paper, we explore the human-like memory-based reasoning on long sequences, where we compress the input stream ${\bf X}$ into a fixed-size memory ${\bf M} = \{{\bf m}_k\}_{k=1}^K$ with $K$ memory slots and then infer the answer ${\bf A}$ for any relevant query ${\bf Q}$ by ${\bf A} = {\mathcal R}({\bf M}, {\bf Q})$. Here we only need to store the compressed memory ${\bf M}$, which can be updated in real-time and reused for a series of queries. Since the slot number $K$ in the memory is irrelevant to the input length $|X|$, this setting only requires $O(1)$ storage space rather than $O(|X|)$ in directly reasoning model ${\mathcal T}({\bf X},{\bf Q})$.

As shown in Figure~\ref{fig:framework}, we apply a rehearsal memory machine ${\mathcal G}_{\Theta}({\bf X})$ to compress the input stream ${\bf X}$ into rehearsal memory ${\bf M} = \{{\bf m}_k\}_{k=1}^K$ with $K$ memory slots. 
\emph{During the training stage}, we simultaneously develop self-supervised rehearsal training and task-specific reasoning training based on the memory $M$. 
For self-supervised rehearsal training, we develop a rehearsal model ${\mathcal H}_{\xi}({\bf M}, {\bf H})$ to reconstruct the masked history fragments~(recollection task) and distinguish positive history fragments from negative ones~(familiarity task), where ${\bf H}$ means the critical history fragments that are selected by the history sampler ${\mathcal S}_{\Psi}({\bf Q},{\bf X})$.
For task-specific reasoning training, we develop the task-specific reason model ${\mathcal R}_{\Omega}({\bf M}, {\bf Q})$ to answer the given query ${\bf Q}$.
\emph{ During the testing stage}, we maintain the rehearsal memory ${\bf M} = {\mathcal G}_{\Theta}({\bf X})$ from the stream ${\bf X}$ and then infer the answer ${\bf A}$ for any relevant query ${\bf Q}$ by ${\bf A} = {\mathcal R}_{\Omega}({\bf M}, {\bf Q})$, where the rehearsal model ${\mathcal H}_{\xi}({\bf M}, {\bf H})$ and the history sampler ${\mathcal S}_{\Psi}({\bf Q},{\bf X})$ are no longer needed.

\subsection{Rehearsal Memory Machine}
We deal with the input stream ${\bf X}$ from the segment level rather than item level, i.e., we cut the input sequence into fixed-length segments and memorize them into the rehearsal memory segment-by-segment. Compared to existing MANNs~\citep{graves2014neural,graves2016hybrid}, which store the input stream item-by-item orderly with a RNN-based controller, our segment-level memorization can further capture the bi-directional context of each item and improve the modeling efficiency. We denote the $t$-th segment as ${\bf X}^t = \{{\bf x}^t_n\}_{n=1}^N$ with $N$ items and the current memory as ${\bf M}^t = \{{\bf m}^t_k\}_{k=1}^K$, where we have recorded $t$-1 segments in ${\bf M}^t$. The ${\bf x}^t_n$ and ${\bf m}^t_k$ have the same dimension $d_{x}$.

We first model the $t$-th segment by a Transformer encoder~\citep{vaswani2017attention} and obtain the sequence features ${\bf F}^t = \{{\bf f}^t_n\}_{n=1}^N$ with dimension $d_{x}$.
After it, we apply a memory update module to write ${\bf F}^t$ into ${\bf M}^t$. We apply a slot-to-item attention to align the sequence features to slot features in the current memory ${\bf M}^t$, and then develop the gate-based update. Concretely, we first calculate the slot-to-item attention matrix where each element means the relevance of a slot-item pair, and then learn aligned features ${\bf L}^t = \{{\bf l}^t_k\}_{k=1}^K$ for each slot, given by
\begin{eqnarray}
\begin{aligned}
& \alpha^t_{kn} = {\bf w}_a^{\top}{\rm tanh}({\bf W}_1^a{\bf m}^t_k + {\bf W}_2^a{\bf f}^t_n + {\bf b}^a), \\  
& {\hat \alpha^t_{kn}} = \frac{{\rm exp}(\alpha^t_{kn})}{\sum_{j=1}^{K} {\rm exp}(\alpha^t_{jn})}, \ {\bf l}^t_k = \sum_{n=1}^{N} {\hat \alpha^t_{kn}} {\bf f}^t_{n},
\end{aligned}
\end{eqnarray}
where ${\bf W}_1^a \in \mathbb{R}^{d_{model} \times d_x}$ , ${\bf W}_2^a  \in \mathbb{R}^{d_{model} \times d_x} $ and ${\bf b}^a  \in \mathbb{R}^{d_{model}} $ are the projection matrices and bias. ${\bf w}_a^{\top}$ is the row vector.
Next, the $k$-th slot feature ${\bf m}^t_{k}$ is updated with its aligned feature ${\bf l}^t_{k}$ based on a GRU unit with ${d_x}$-d hidden states, given by
\begin{eqnarray}
{\bf m}^{t+1}_{k} = {\rm GRU}({\bf m}^t_{k}, {\bf l}^t_{k}),
\end{eqnarray}
where ${\bf l}^t_{k}$ is the current input of the GRU unit and ${\bf m}^{t}_{k}$ is the hidden state at the $t$-th step. And ${\bf m}^{t+1}_{k}$ is the new slot feature after the gate-based update. After memorizing $T$ segments, we can obtain rehearsal memory ${\bf M}^{T+1}$ and we denote it by ${\bf M}$ for convenience.

\subsection{Self-Supervised Rehearsal Training}
Based on the maintained memory $M$, we apply the memory rehearsal technique to enhance long-sequence memorization. We first design the self-supervised recollection and familiarity tasks to solve \emph{how to rehearse}, which can alleviate the gradual forgetting of early information. We next adopt a history sampler to decide \emph{what to rehearse}, which guides the memory to remember critical task-relevant clues.

\subsubsection{How to Rehearse: Self-Supervised Recollection and Familiarity Tasks}
We design the rehearsal model ${\mathcal H}_{\xi}({\bf M}, {\bf H})$ with recollection and familiarity tasks.
The recollection task reconstructs the masked positive history fragments to enable the memory to relive past episodes.
And the familiarity task tries to distinguish positive history fragments from negative ones for making the memory recognize experienced events.

First, we apply an independent history sampler to select the $B$ segments from the input stream as the history fragment set ${\bf H} = \{H^b\}_{b=1}^B$, which is illustrated in the next section. Each fragment $H^b = \{{h}^b_1, {h}^b_2, \cdots, {h}^b_{N}\}$ contains $N$ items and each item ${h}_{*}$ corresponds to a feature ${\bf x}_{*}$.
For the $b$-th fragment, we randomly mask 50\% of items in the fragment and add an especial item [cls] at the beginning to obtain the masked positive history fragment ${H}^{b+} = \{{h}^{b+}_{[cls]}, {h}^{b+}_1,  {h}^{b+}_{[{m}_1]}, \cdots, {h}^{b+}_{N}\}$, where ${h}^{b+}_{[{m}_1]}$ means the first masked item. 
In order to guarantee that the model ${\mathcal H}_{\xi}({\bf M}, {\bf H})$ reconstructs the masked fragment by utilizing the maintained memory ${\bf M}$ rather than only relying on fragment context, we set the mask ratio to 50\% instead of 15\% in BERT~\citep{devlin2019bert}.
Moreover, we construct the masked negative history fragment ${H}^{b-} = \{{h}^{b-}_{[cls]}, {h}^{b-}_1,  {h}^{b-}_{[{m}_1]}, \cdots, {h}^{b-}_{N}\}$ by replacing 50\% of unmasked items in the positive fragment, where the replacement items are sampled from other input stream to make the negative fragment distinguishable.
Here we construct the positive fragment set ${\bf H}^+ = \{H^{b+}\}_{b=1}^B$ and corresponding negative fragment set ${\bf H}^- = \{H^{b-}\}_{b=1}^B$ from the original fragment set ${\bf H}$. 
Next, we adopt a bidirectional Transformer decoder~\citep{vaswani2017attention} without the future masking to model each history fragment ${H}^{b+}/{H}^{b-}$. In the decoder, each history item can interact with all other items in the fragment. 
The rehearsal memory ${\bf M}$ is input to the ``encoder-decoder multi-head attention su{b-}layer'' in each decoder layer, where the queries come from the previous decoder layer and the memory slots are regarded as the keys and values. This allows each item in the decoder to attend over all slot features in the memory ${\bf M}$ .
Finally, we obtain the features $\{{\bf r}^{b+/b-}_{[cls]}, {\bf r}^{b+/b-}_1,  {\bf r}^{b+/b-}_{[{m}_1]}, \cdots, {\bf r}^{b+/b-}_{N}\}$ where each ${\bf r}^{b+/b-}_{*}$ has the dimension $d_{x}$.

{\bf Recollection Task.} We first predict the masked items of positive history fragments to build the item-level reconstruction for the recollection task. Considering there are too many item types, we apply the contrastive training~\cite{he2019momentum,chen2020simple,zhang2020counterfactual} based on the ground truth and other sampled items. For the $N/2$ masked items, we compute the recollection loss for the $b$-th fragment by 
\begin{equation}
\begin{aligned}
&{\mathcal L}_{i}^b  = {\rm log}\frac{ {\rm exp}({\bf r}^{b+}_{[{m}_i]} \cdot {\bf y}_i)  }{{\rm exp}({\bf r}^{b+}_{[{m}_i]} \cdot {\bf y}_i)  + \sum_{j=1}^{J} {\rm exp}({\bf r}^{b+}_{[{m}_i]} \cdot {\bf y}_j) }, \\
&{\mathcal L}_{rec}^b  = -\frac{2}{N}\sum_{i=1}^{N/2}{\mathcal L}_{i}^b
\end{aligned}
\end{equation}
where ${\bf y}_i \in \mathbb{R}^{d_x}$ is the feature of ground truth of the $i$-th masked item,  ${\bf y}_j \in \mathbb{R}^{d_x} $ is the feature of sampled items and ${\bf r}^{b+}_{[{m}_i]} \cdot {\bf y}_*$ is the inner product of two features.

{\bf Familiarity Task.} Next, we predict whether the masked history fragment ever appears in the current input stream, i.e. distinguish positive history fragments from negative ones. This training objective makes the memory learn the ability of recognizing experienced events. Concretely, we project each feature ${\bf r}^{b+/b-}_{[cls]}$ into a confident score ${s}^{b+/b-} \in (0,1)$ by a linear layer with the sigmoid activation, and calculate the familiarity loss by
\begin{equation}
{\mathcal L}_{fam}^b  = - {\rm log}(s^{b+}) + {\rm log}(1-s^{b-}),
\end{equation}
where ${\mathcal L}_{fam}^b$ is the familiarity loss for the $b$-th pair of positive and negative fragments in ${\bf H}^+/{\bf H}^-$.

\subsubsection{What to Rehearse: History Sampler}
Existing MANNs~\citep{graves2014neural,graves2016hybrid} often have no assumptions on what information needs to be remembered the most. But due to the limited capacity of the memory, it is crucial to distinguish what contents are important for subsequent inference and pay more attention to them during memorization.  Thus, we further train a history sampler ${\mathcal S}_{\Psi}({\bf Q},{\bf X})$ to select informative history fragments ${\bf H}$ for self-supervised rehearsal training, which is independent to the memory machine ${\bf M} = {\mathcal G}_{\Theta}({\bf X})$.
In knowledge distillation~\citep{hinton2015distilling}, the teacher model can access the privileged information and transfer the knowledge to the student model. Similar to it, our history sampler, which is the teacher and can access raw contents ${\bf X}$ while answering the query ${\bf Q}$, learns the ability of distinguishing task-relevant important fragments and guides the rehearsal memory (i.e., the student) to remember critical clues.

Concretely, we independently train a directly reasoning model as the history sampler ${\mathcal S}_{\Psi}({\bf Q},{\bf X})$, which can access raw contents ${\bf X}$ while answering the query ${\bf Q}$. We first cut the entire input ${\bf X}$ into $C$ history fragments $\{H_c\}_{c=1}^C$ just like the rehearsal memory machine, where each fragment contains $N$ items. Next, we obtain the fragment features $\{{\bf h}_c\}_{c=1}^C$ by averaging the item features in each fragment. After it, we develop the attention-based reasoning for the query ${\bf Q}$ on these fragment features. The query feature ${\bf q} \in \mathbb{R}^{d_{model}}$ is modeled by the task-specific encoder in different downstream tasks, which is introduced in Section~\ref{app.1} of the supplementary material. 
Given the query feature ${\bf q}$ and fragment features $\{{\bf h}_c\}_{c=1}^C$, we conduct the attention method to aggregate query-relevant clues from fragments, given by
\begin{eqnarray}
\begin{aligned}
& \beta_{c} = {\bf w}_h^{\top}{\rm tanh}({\bf W}_1^h{\bf q} + {\bf W}_2^h{\bf h}_c + {\bf b}^h), \\  
& {\hat \beta_c} = \frac{{\rm exp}(\beta_c)}{\sum_{j=1}^{C} {\rm exp}(\beta_j)}, \ {\bf e} = \sum_{c=1}^{C} {\hat \beta_c} {\bf h}_{c}, 
\end{aligned}
\end{eqnarray}
where ${\bf W}_1^h \in \mathbb{R}^{d_{model} \times d_{model}}$ , ${\bf W}_2^h  \in \mathbb{R}^{d_{model} \times d_x} $ and ${\bf b}^h  \in \mathbb{R}^{d_{model}} $ are the projection matrices and bias. And ${\bf w}_h^{\top}$ is the row vector.
We then obtain the reasoning feature ${\bf a} = [{\bf e}; {\bf q}] $ by concatenating the query and query-relevant fragment features, and design the final reasoning layer for different tasks, shown in Section~\ref{app.1} of the supplementary material. After independent training, the attention weights $\{\beta_{c}\}_{c=1}^C$ can be regarded as the importance score of each fragment for the query ${\bf Q}$. Thus, we can sample the vital history fragments with high attention weights for each $({\bf X}, {\bf Q})$ pair. Specifically, to guarantee the sampled fragments appear in the entire input stream, we select $B/2$ fragments from $\{H_c\}_{c=1}^{C/2}$ with large weights and choose another $B/2$ fragments from $\{H_c\}_{c=C/2}^{C}$ to constitute the fragment set ${\bf H} = \{H^b\}_{b=1}^B $. The final recollection loss ${\mathcal L}_{rec}$ and familiarity loss  ${\mathcal L}_{fam}$ are computed by
\begin{equation}
{\mathcal L}_{rec}  =  \mathbb{E}_{b \sim {\mathcal S}_{\Psi}} [{\mathcal L}_{rec}^b] , \ \  {\mathcal L}_{fam}  =  \mathbb{E}_{b \sim {\mathcal S}_{\Psi}} [{\mathcal L}_{fam}^b].
\end{equation}

\subsection{Task-Specific Reasoning Training}
Besides self-supervised rehearsal training, we simultaneously develop task-specific reasoning training. For several downstream tasks, we propose different task-specific reason model ${\mathcal R}_{\Omega}({\bf M}, {\bf Q})$ based on the memory ${\bf M}$. Here we adopt the simple and mature components in the reason model for a fair comparison. The details are introduced in Section~\ref{app.1} of the supplementary material. 
Briefly, we first learn the query representation ${\bf q}$ by a task-specific encoder and then perform the multi-hop attention-based reasoning. Finally, we obtain the reason loss ${\mathcal L}_{r}$ from ${\mathcal R}_{\Omega}({\bf M}, {\bf Q})$. 

Eventually, we combine the rehearsal and reason losses to train our model, given by
\begin{eqnarray}
{\mathcal L}_{rm} = {\lambda}_1 {\mathcal L}_{rec} +  {\lambda}_2 {\mathcal L}_{fam} + {\lambda}_3 {\mathcal L}_{r},
\end{eqnarray}
where $\lambda_1$, $\lambda_2$ and $\lambda_3$ are applied to adjust the balance of three losses.

\section{Experiments}
In this section, we first verify our rehearsal memory on the widely-used short-sequence reasoning task bAbI. Next, we mainly compare our approach with diverse baselines on several long-sequence reasoning tasks.
We then perform ablation studies on the memory rehearsal techniques and analyze the impact of crucial hyper-parameters.

\subsection{Experiment Setting}
{\bf Model Setting}. We first introduce the common model settings for all downstream tasks. We set the layer number of the Transformer encoder and bi-directional Transformer decoder to 3. The head number in Multi-Head Attention is set to 4. We set $\lambda_1$, $\lambda_2$ and $\lambda_3$ to 1.0, 0.5 and 1.0, respectively. The number $B$ of history fragments is set to 6.
During training, we apply an Adam optimizer~\citep{duchi2011adaptive} to minimize the multi-task loss ${\mathcal L}_{rm}$, where the initial learning rate is set to 0.001.

{\bf Baseline.} We compare our rehearsal memory with the directly reasoning methods and the memory-based reasoning approaches. The directly reasoning baselines are different in downstream tasks and the memory-based baselines mainly are DNC~\citep{graves2016hybrid}, NUTM~\citep{le2019neural},  DMSDNC~\citep{park2020distributed}, STM~\citep{le2020self} and Compressive Transformer (CT)~\cite{rae2019compressive}.
For a fair comparison, we modify the reasoning module of memory-based baselines to be consistent with our rehearsal memory, i.e. we conduct multi-hop attention-based reasoning based on the built memory. And the number of memory slots in these baselines is also set to K. Besides, we set the core number of NUTM to 4, the query number of STM to 8 and the memory block number of DMSDNC to 2. As for CT, the layer number of the Transformer is set to 3 as our rehearsal memory and the compression rate is set 5.

\subsection{Rehearsal Memory on Short-Sequence Reasoning}
The bAbI dataset~\citep{weston2015towards} is a synthetic text question answering benchmark and widely applied to evaluate the memorization and reasoning performance of MANNs. This dataset contains 20 reasoning tasks and requires to be solved with one common model. Although most of these tasks only give short-sequence text input (less than 100 words) and existing methods~\citep{park2020distributed,le2020self} have solved these tasks well, we still compare our rehearsal memory with other memory-based baselines to verify the short-sequence reasoning performance. 
We set the $d_x$ and $d_{model}$ to 128. The number K of memory slots is set to 20. And we naturally take each sentence in input texts as a segment and the maximum length N of segments is set to 15. Due to limited word types in this dataset, we sample all other words as negative items in ${\mathcal L}_{rec}$.

\begin{table}[t]
\centering
    \begin{adjustbox}{max width = 1.0 \textwidth}
    \begin{threeparttable}
    \caption{Performance Comparisons for Synthetical bAbI Task: mean $\pm$ std. and best error over 10 runs.}
    \label{table:babi}
        \begin{tabular}{c|c|c}
            \toprule
            {Method} & mean $\pm$ std error & best error\\
            \midrule
                DNC & 16.7 $\pm$ 7.6& 3.8\\
                NUTM& 5.6 $\pm$ 1.9& 3.3\\
                {DMSDNC}&  1.53 $\pm$ 1.33 &0.16  \\
                {STM} &  0.39 $\pm$ 0.18& 0.15\\
                CT & 0.81 $\pm$ 0.26 & 0.34 \\
            \midrule
				RM &  {\bf 0.33 $\pm$ 0.15} & {\bf 0.12}\\            
			\bottomrule

        \end{tabular}
      \end{threeparttable}
      \end{adjustbox}
\end{table}

The results are summarized in Table~\ref{table:babi}. The RM model solves these bAbI tasks with near-zero error and outperforms existing baselines at the mean and best error rate of 10 runs, verifying that our RM method can conduct effective memorization and reasoning on short-sequence tasks.
In these methods, DNC, NUTM and DMSDNC model the input contents word-by-word. STM processes input texts as a sentence-level sequence. And CT and RM methods cut the input texts into sentences for modeling. From the results, we can find STM, CT and RM  achieve lower error rates than other baselines, suggesting the importance of sentence-level modeling. Considering the bAbI tasks are close to being solved, we design another difficult synthetic task in Section~\ref{sec:syn} of the supplementary material to evaluate our RM model.

\subsection{Rehearsal Memory on Long-Sequence Reasoning}
We then compare our approach with diverse baselines on several long-sequence reasoning tasks.

\subsubsection{Long-Sequence Text Question Answering}
We apply the NarrativeQA dataset~\citep{kovcisky2018narrativeqa} with long input contents for long-sequence text question answering. This dataset contains 1,572 stories and corresponding summaries generated by humans, where each summary contains more than 600 tokens on average. And there are 46,765 questions in total. We adopt the multi-choice form to answer the given question based on a summary, where other answers for questions associated with the same summary are regarded as answer candidates. 
We compute the mean reciprocal rank (MRR) as the metric, i.e., the rank of the correct answer among candidates. 
Besides the memory-based methods, we adopt directly reasoning model {\bf AS Reader}~\citep{kadlec2016text} and {\bf E2E-MN}~\citep{sukhbaatar2015end} as baselines.
The AS Reader applies a pointer network to generate the answer and E2E-MN employs the end-to-end memory network to conduct multi-hop reasoning. 
For our rehearsal memory, we set the $d_x$ and $d_{model}$ to 256. The number K of memory slots is set to 20. We naturally take each sentence in summaries as a segment and the maximum length N of segments is set to 20. And we sample all other words as negative items in ${\mathcal L}_{rec}$.

\begin{table}[t]
\centering
    \begin{adjustbox}{max width = 1.0\textwidth}
    \begin{threeparttable}
    \caption{Performance Comparisons for Long-Sequence Text Question Answering on NarrativeQA.}
    \label{table:narrative}
        \begin{tabular}{c|c|c|c}
            \toprule
            {Method} &{Setting}& Val MRR & Test MRR \\
            \midrule
            AS Reader & Directly &26.9 & 25.9\\ 
            E2E-MN & Directly & 29.1& 28.6\\
            \midrule
            \midrule
                DNC &Memory-Based& 25.8& 25.2\\
                NUTM&Memory-Based& 27.7& 27.2\\
                {DMSDNC}& {Memory-Based} & {28.1} &{27.5}  \\
                {STM} & {Memory-Based}& {27.2}& {26.7} \\
                CT & {Memory-Based} & 28.7 & 28.3 \\
            \midrule
                RM & Memory-Based& {\bf 29.4} & {\bf 28.7}\\
            \bottomrule

        \end{tabular}
      \end{threeparttable}
      \end{adjustbox}

\end{table}

We report the results in Table~\ref{table:narrative}. Our RM method obtains the best performance among memory-based approaches, which demonstrates our self-supervised rehearsal training with the history sampler can effectively enhance long-sequence memorization and reasoning. Further, the RM model slightly outperforms early directly reasoning methods AS Reader and E2E-MN, showing the ability of efficient reasoning on long sequences with limited storage resources.

\subsubsection{Long-Sequence Video Question Answering}
The ActivityNet-QA dataset~\citep{yu2019activitynet} contains 5,800 videos from the ActivityNet~\citep{caba2015activitynet}.  The average video duration of this dataset is about 180s and is the longest in VQA datasets. 
We compare our method with four directly reasoning baselines, including three basic models {\bf E-VQA}, {\bf E-MN}, {\bf E-SA} from \citep{yu2019activitynet} and the SOTA model {\bf HCRN}~\citep{le2020hierarchical}. 
For our rehearsal memory, we set the $d_x$ and $d_{model}$ to 256. The number K of memory slots and length N of segments are both set to 20. And in ${\mathcal L}_{rec}$, we select 30 other frame features from the video as the sampled items.

As shown in Table~\ref{table:vqa}, the RM method obtains a better performance than other memory-based baselines. Compared to the best baseline CT, our RM model further achieves the 0.9\% absolute improvement, showing the effectiveness of our model designs and self-supervised rehearsal training. Moreover, the RM method outperforms the basic directly reasoning baselines E-VQA, E-MN and E-SA, but slightly worse than the SOTA method HCRN. This suggests our rehearsal memory can reduce the gap between memory-based and directly reasoning paradigms.

\begin{table}[t]
\centering
        \begin{adjustbox}{max width = 1\textwidth}
    \begin{threeparttable}
    \caption{Performance Comparisons for Long-Term Video Question Answering on ActivityNet-QA.}
    \label{table:vqa}
        \begin{tabular}{c|c|c}
            \toprule
            {Method} &{Setting} & Accuracy  \\
            \midrule
            E-VQA & Directly&25.2\\
            E-MN & Directly&27.9\\ 
            E-SA & Directly&31.8\\ 
            HCRN & Directly&37.6\\
            \midrule
            \midrule
            DNC &Memory-Based&30.3\\
            NUTM&Memory-Based&33.1\\
            DMSDNC&Memory-Based&32.4\\
            STM&Memory-Based&33.7\\
            CT&Memory-Based&35.4\\

            \midrule
            RM & Memory-Based&{\bf 36.3}\\
            \bottomrule

        \end{tabular}
      \end{threeparttable}
      \end{adjustbox}

\end{table}

\begin{table}[t]
\centering
    \begin{adjustbox}{max width = 1\textwidth}
    \begin{threeparttable}
    \caption{Performance Comparisons for Lifelong Sequence Recommendation on XLong.}
    \label{table:xlong}
        \begin{tabular}{c|c|c}
            \toprule
            {Method} &{Setting}&AUC  \\
            \midrule
            GRU4REC&Directly&0.8702\\
            Caser&Directly&0.8390\\ 
            RUM & Directly&0.8649\\
            DIEN&Directly&0.8793\\  
            \midrule
            \midrule
                HPMN&Memory-Based&0.8645\\
                MIMN&Memory-Based&0.8731\\
            \midrule
                RM &Memory-Based&{\bf 0.8817}\\
            \bottomrule

        \end{tabular}
      \end{threeparttable}
    \end{adjustbox}

\end{table}

\subsubsection{Lifelong Sequence Recommendation}
The lifelong sequence recommendation~\citep{ren2019lifelong} aims to predict whether the user will click a given item based on long sequences, thus it can be regarded as a long-sequence reasoning task. The XLong dataset~\citep{ren2019lifelong} is sampled from the click logs on Alibaba. The length of historical behavior sequences in this dataset is 1000. We compare our method with four directly reasoning methods {\bf GRU4REC}~\citep{hidasi2015session}, {\bf Caser}~\citep{tang2018personalized}, {\bf DIEN}~\citep{zhou2019deep}, {\bf RUM}~\citep{chen2018sequential} and two memory-based methods {\bf HPMN}~\citep{ren2019lifelong} and {\bf MIMN}~\citep{pi2019practice}, where the HPMN method builds the memory by hierarchical RNNs and the MIMN method introduces a write-read memory as in~\citep{graves2014neural}.
For our rehearsal memory, we set the $d_x$ and $d_{model}$ to 64. The number K of memory slots and length N of segments are both set to 20. 
And in ${\mathcal L}_{rec}$, we select 200 items from the large item set as the sampled items.

The results are shown in Table~\ref{table:xlong}. our RM method not only outperforms other memory-based approaches, but also achieves better performance than directly reasoning baselines. This is because our rehearsal memory can aggregate and organize the long-term interests from user behavior sequences and these interests can be activated during next-item prediction. But the directly reasoning approaches may fail to learn such informative interest representations.

\begin{table}[t]
\centering
        \begin{adjustbox}{max width = 1\textwidth}
    \begin{threeparttable}
    \caption{Ablation Results about the Rehearsal Losses and History Sampler.}
    \label{table:ablation}
        \begin{tabular}{ccccc}
            \toprule
            \multirow{2}{*}{Method}&\multicolumn{2}{c}{NarrativeQA} & ActNet-QA & XLong\\
            \cmidrule(lr){2-3} \cmidrule(lr){4-4} \cmidrule(lr){5-5}
               &   Val&   Test &  Acc. & AUC \\
            \midrule
            w/o. rehearsal & 27.9&27.5&24.6&0.8745\\
            only ${\mathcal L}_{rec}$& 29.2&28.6&36.0&0.8802\\
            only ${\mathcal L}_{fam}$& 28.6&28.1&35.4&0.8776\\
            random sampler &28.7&28.3&35.7&0.8813\\
            \midrule
            Full & {\bf 29.4} & {\bf 28.7} &{\bf 36.3}&{\bf 0.8817}\\
            \bottomrule

        \end{tabular}
      \end{threeparttable}
      \end{adjustbox}

\end{table}

\begin{table}[t]
\centering
        \begin{adjustbox}{max width = 0.9\textwidth}

      \begin{threeparttable}

\caption{Ablation Results about the Transformer Encoder and Mask Ratio.}\label{table:ablation2}

        \begin{tabular}{ccccc}
            \toprule
            \multirow{2}{*}{Method}&\multicolumn{2}{c}{NarrativeQA} & ActNet-QA & XLong\\
            \cmidrule(lr){2-3} \cmidrule(lr){4-4} \cmidrule(lr){5-5}
               &   Val&   Test &  Acc. & AUC \\
            \midrule
            GRU Encoder & 29.1 &28.3& 35.9& 0.8789\\
            15\% Mask Ratio &  29.0 & 28.5&36.0&0.8809\\
            \midrule
            Full & {\bf 29.4} & {\bf 28.7} &{\bf 36.3}&{\bf 0.8817}\\
            \bottomrule

        \end{tabular}
      \end{threeparttable}
      \end{adjustbox}


\end{table}

\subsection{Ablation Study for Self-Supervised Rehearsal}
We next perform ablation studies on the self-supervised rehearsal losses and history sampler. 
Concretely, we first completely discard the self-supervised rehearsal training to produce the ablation model RM (w/o. rehearsal). We then remove the recollection or familiarity loss to produce two ablation models RM (only ${\mathcal L}_{fam}$) and RM (only ${\mathcal L}_{rec}$), where the history sampler is still retained. 
Next, we replace the independent history sampler with a random sampler to generate the ablation model RM (random sampler), which randomly selects $B$ history fragments from the input stream for rehearsal training.

We conduct the ablation experiments on NarrativeQA, ActivityNet-QA and XLong datasets. The results are reported in Table~\ref{table:ablation}. We can find the full model outperforms the model RM (w/o. rehearsal), demonstrating the self-supervised rehearsal training with the history sampler can further boost the long-sequence memorization and reasoning ability of rehearsal memory. 
Further, the full model has better performance than RM (only ${\mathcal L}_{fam}$) and RM (only ${\mathcal L}_{rec}$) on all metrics, which illustrates two rehearsal tasks are both helpful for alleviating the issue of gradual forgetting. 
And RM (only ${\mathcal L}_{rec}$) achieves better results than RM (only ${\mathcal L}_{fam}$), showing the recollection task that enables the memory to relive past episode is more important for rehearsal training.
Moreover, the ablation model RM (random sampler) has the performance degradation than the full model. This fact indicates it is critical to select informative history fragments for rehearsal training. By the guidance of the history sampler, the rehearsal memory can remember task-relevant clues for subsequent reasoning.

\subsection{Ablation Study for Model Setting}
In this section, we conduct ablation study aboout the model settings. 
Existing MANNs often use a RNN as their controller, but we apply a Transformer encoder to improve the sequential modeling ability of RM.
Thus, we replace the Transformer encoder in the rehearsal memory machine with a bi-directional GRU encoder.
As shown in Table~\ref{table:ablation2}, the full model achieves better performance than the model with the GRU Encoder, verifying the effectiveness of the Transformer encoder.

For the masked history fragments, we set the mask ratio to 50\%  instead of 15\% in BERT. We compare the results of two mask ratios in Table~\ref{table:ablation2}. We can find the full model with the 50\% mask ratio outperforms the one with the 15\% mask ratio. This fact suggests that the large mask ratio makes the rehearsal model ${\mathcal H}_{\xi}({\bf M}, {\bf H})$  utilize the maintained memory ${\bf M}$ than only relying on fragment context, and is beneficial for the self-supervised rehearsal training of RM.

\begin{figure}[t]
\centering
    \subfigure[NarrativeQA]{
        \includegraphics[width=0.47\columnwidth]{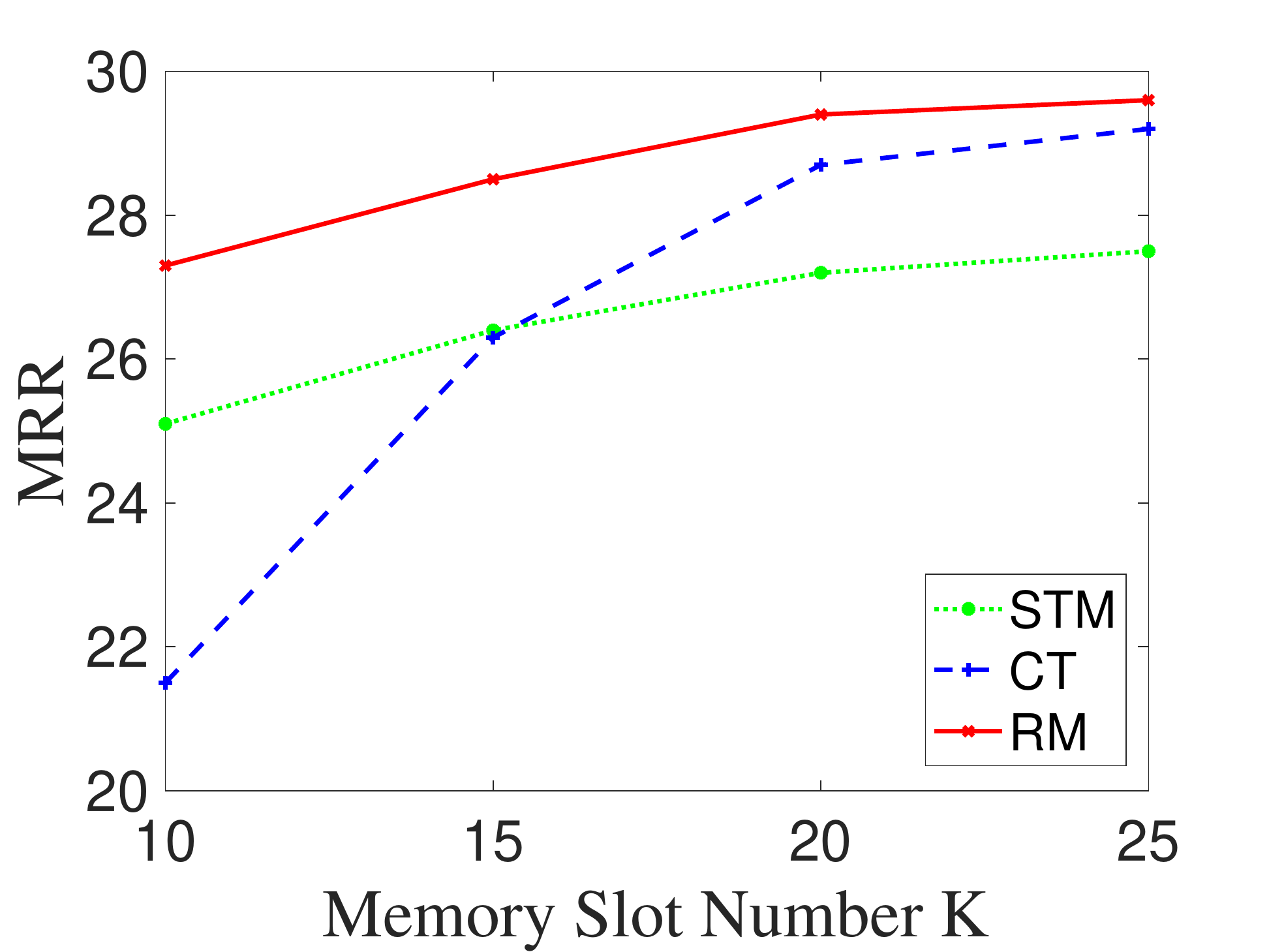}
    }
    \subfigure[ActivityNet-QA]{
        \includegraphics[width=0.47\columnwidth]{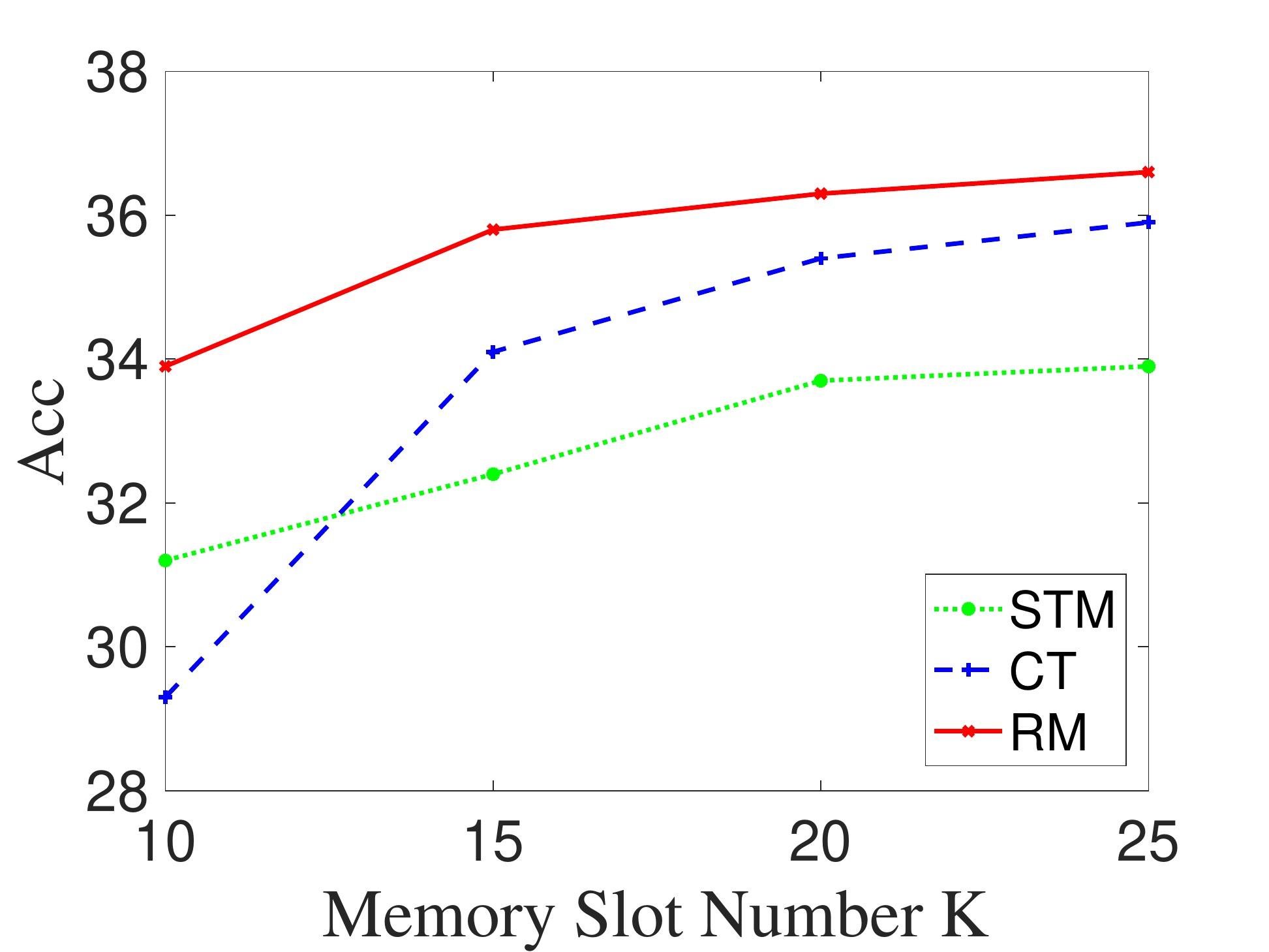}
    }

\caption{Effect of the Memory Slot Number K.}\label{fig:hyperslot}
\end{figure}

\begin{figure}[t]
\centering

    \includegraphics[width=0.6\columnwidth]{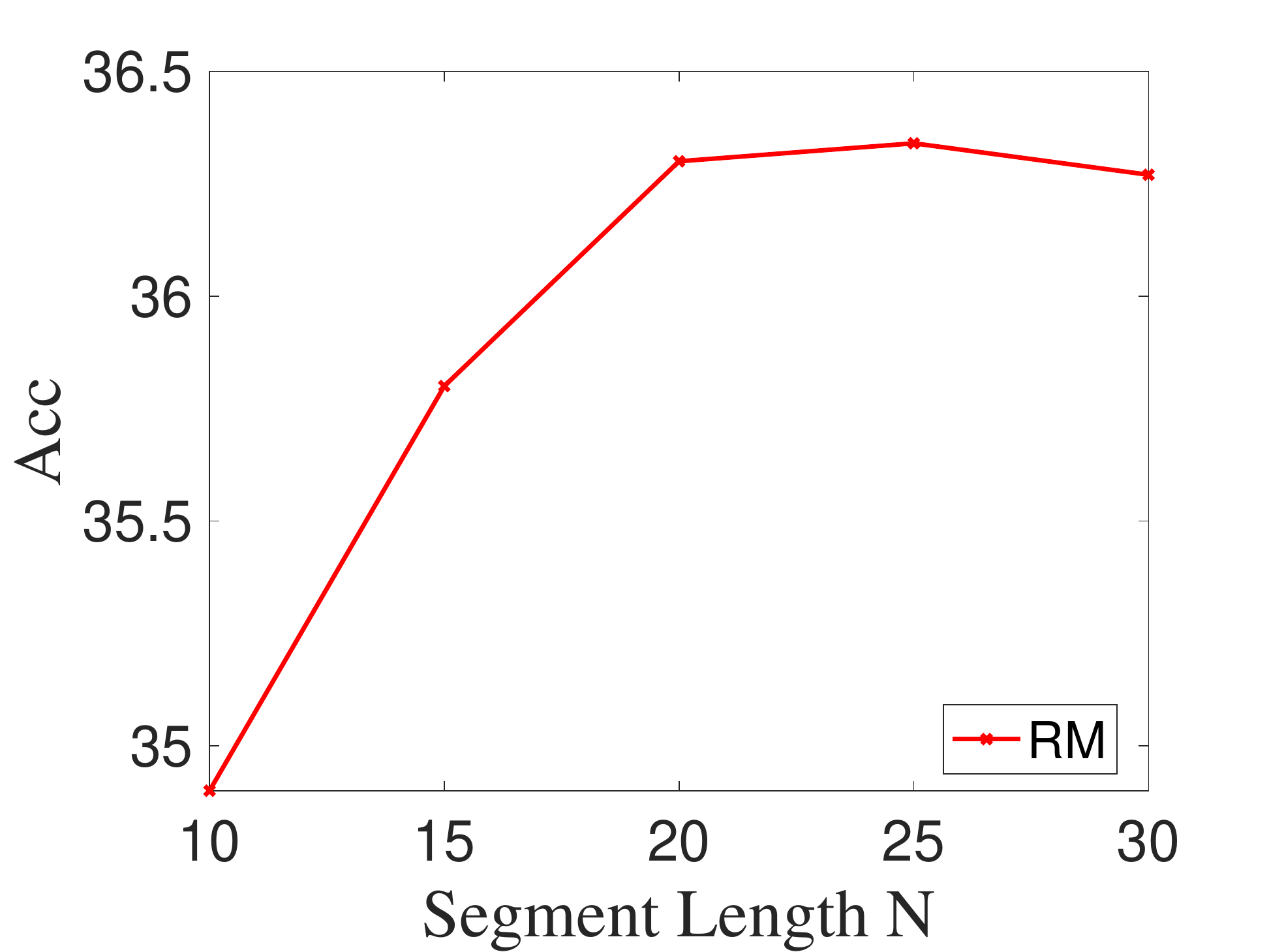}
\caption{Effect of the Segment Length N on the ActivityNet-QA dataset.}\label{fig:hypersegment}

\end{figure}

\subsection{Hyper-Parameters Analysis}
We then explore the effect of two crucial hyper-parameters: the memory slot number $K$ and the segment length $N$. 
We first set the slot number $K$ to [10, 15, 20, 25] and compare our RM method with two baselines STM and CT on the NarrativeQA and ActivityNet-QA datasets. We display the results in Figure~\ref{fig:hyperslot}. 
We note that the performance of all three methods gradually improves with the increase of slot number and slowly reaches the bottleneck. When the number of memory slots exceeds 20, more slots can not bring much improvement. By comparison, we can find our RM method achieves the best performance on different slot numbers, verifying the effectiveness and stability of our rehearsal memory. Moreover, the performance of CT is terrible when the slot number is too few. This is because the CT method implements the compressed memory by a FIFO queue and completely forgets the contents beyond the queue, i.e., its memorization range severely depends on the slot number.

We then set the segment length $N$ to [10, 15, 20, 25, 30] and report the results on the ActivityNet-QA dataset in Figure~\ref{fig:hypersegment}.
We can find that when the segment length is set to 10, the RM method achieves poor results and the performance is relatively stable when the segment length changes between 20 and 30. This is because when the segment is too short, important evidence may be scattered in different segments, and the model cannot effectively capture the evidence and infer the answer.

\section{Conclusions}
In this paper, we propose the self-supervised rehearsal to enhance long-sequence memorization. 
We design the self-supervised recollection and familiarity tasks to alleviate the gradual forgetting of early information. Further, we adopt a history sampler to guide the memory to remember critical information.
Extensive experiments on a series of downstream tasks verify the performance of our method.
For future work, we will further explore the property of rehearsal memory.

\section*{Acknowledgments}
This work is supported by the National Key R\&D Program of China under Grant No. 2018AAA0100603. This research is supported by the National Natural Science Foundation of China under Grant No.61836002 and No.62072397, and the Zhejiang Natural Science Foundation LR19F020006.





\bibliography{all}

\begin{thebibliography}{39}
\providecommand{\natexlab}[1]{#1}
\providecommand{\url}[1]{\texttt{#1}}
\expandafter\ifx\csname urlstyle\endcsname\relax
  \providecommand{\doi}[1]{doi: #1}\else
  \providecommand{\doi}{doi: \begingroup \urlstyle{rm}\Url}\fi

\bibitem[Baddeley(1992)]{baddeley1992working}
Baddeley, A.
\newblock Working memory.
\newblock \emph{Science}, 255\penalty0 (5044):\penalty0 556--559, 1992.

\bibitem[Caba~Heilbron et~al.(2015)Caba~Heilbron, Escorcia, Ghanem, and
  Carlos~Niebles]{caba2015activitynet}
Caba~Heilbron, F., Escorcia, V., Ghanem, B., and Carlos~Niebles, J.
\newblock Activitynet: A large-scale video benchmark for human activity
  understanding.
\newblock In \emph{Proceedings of the IEEE Conference on Computer Vision and
  Pattern Recognition}, pp.\  961--970, 2015.

\bibitem[Chen et~al.(2020)Chen, Kornblith, Norouzi, and Hinton]{chen2020simple}
Chen, T., Kornblith, S., Norouzi, M., and Hinton, G.
\newblock A simple framework for contrastive learning of visual
  representations.
\newblock \emph{arXiv preprint arXiv:2002.05709}, 2020.

\bibitem[Chen et~al.(2018)Chen, Xu, Zhang, Tang, Cao, Qin, and
  Zha]{chen2018sequential}
Chen, X., Xu, H., Zhang, Y., Tang, J., Cao, Y., Qin, Z., and Zha, H.
\newblock Sequential recommendation with user memory networks.
\newblock In \emph{Proceedings of the eleventh ACM international conference on
  web search and data mining}, pp.\  108--116, 2018.

\bibitem[Csord{\'a}s \& Schmidhuber(2019)Csord{\'a}s and
  Schmidhuber]{csordas2019improving}
Csord{\'a}s, R. and Schmidhuber, J.
\newblock Improving differentiable neural computers through memory masking,
  de-allocation, and link distribution sharpness control.
\newblock \emph{arXiv preprint arXiv:1904.10278}, 2019.

\bibitem[Devlin et~al.(2019)Devlin, Chang, Lee, and Toutanova]{devlin2019bert}
Devlin, J., Chang, M.-W., Lee, K., and Toutanova, K.
\newblock Bert: Pre-training of deep bidirectional transformers for language
  understanding.
\newblock In \emph{Proceedings of the Conference on The North American Chapter
  of the Association for Computational Linguistics}, 2019.

\bibitem[Duchi et~al.(2011)Duchi, Hazan, and Singer]{duchi2011adaptive}
Duchi, J., Hazan, E., and Singer, Y.
\newblock Adaptive subgradient methods for online learning and stochastic
  optimization.
\newblock \emph{Journal of Machine Learning Research}, 12\penalty0
  (Jul):\penalty0 2121--2159, 2011.

\bibitem[Graves et~al.(2014)Graves, Wayne, and Danihelka]{graves2014neural}
Graves, A., Wayne, G., and Danihelka, I.
\newblock Neural turing machines.
\newblock \emph{arXiv preprint arXiv:1410.5401}, 2014.

\bibitem[Graves et~al.(2016)Graves, Wayne, Reynolds, Harley, Danihelka,
  Grabska-Barwi{\'n}ska, Colmenarejo, Grefenstette, Ramalho, Agapiou,
  et~al.]{graves2016hybrid}
Graves, A., Wayne, G., Reynolds, M., Harley, T., Danihelka, I.,
  Grabska-Barwi{\'n}ska, A., Colmenarejo, S.~G., Grefenstette, E., Ramalho, T.,
  Agapiou, J., et~al.
\newblock Hybrid computing using a neural network with dynamic external memory.
\newblock \emph{Nature}, 538\penalty0 (7626):\penalty0 471--476, 2016.

\bibitem[He et~al.(2020)He, Fan, Wu, Xie, and Girshick]{he2019momentum}
He, K., Fan, H., Wu, Y., Xie, S., and Girshick, R.
\newblock Momentum contrast for unsupervised visual representation learning.
\newblock In \emph{Proceedings of the IEEE Conference on Computer Vision and
  Pattern Recognition}, 2020.

\bibitem[Hidasi et~al.(2015)Hidasi, Karatzoglou, Baltrunas, and
  Tikk]{hidasi2015session}
Hidasi, B., Karatzoglou, A., Baltrunas, L., and Tikk, D.
\newblock Session-based recommendations with recurrent neural networks.
\newblock \emph{arXiv preprint arXiv:1511.06939}, 2015.

\bibitem[Hinton et~al.(2015)Hinton, Vinyals, and Dean]{hinton2015distilling}
Hinton, G., Vinyals, O., and Dean, J.
\newblock Distilling the knowledge in a neural network.
\newblock \emph{arXiv preprint arXiv:1503.02531}, 2015.

\bibitem[Jin et~al.(2019{\natexlab{a}})Jin, Zhao, Gu, Yu, Xiao, and
  Zhuang]{jin2019multi}
Jin, W., Zhao, Z., Gu, M., Yu, J., Xiao, J., and Zhuang, Y.
\newblock Multi-interaction network with object relation for video question
  answering.
\newblock In \emph{Proceedings of the ACM International Conference on
  Multimedia}, pp.\  1193--1201, 2019{\natexlab{a}}.

\bibitem[Jin et~al.(2019{\natexlab{b}})Jin, Zhao, Gu, Yu, Xiao, and
  Zhuang]{jin2019video}
Jin, W., Zhao, Z., Gu, M., Yu, J., Xiao, J., and Zhuang, Y.
\newblock Video dialog via multi-grained convolutional self-attention context
  networks.
\newblock In \emph{Proceedings of the International ACM SIGIR Conference on
  Research and Development in Information Retrieval}, pp.\  465--474,
  2019{\natexlab{b}}.

\bibitem[Kadlec et~al.(2016)Kadlec, Schmid, Bajgar, and
  Kleindienst]{kadlec2016text}
Kadlec, R., Schmid, M., Bajgar, O., and Kleindienst, J.
\newblock Text understanding with the attention sum reader network.
\newblock \emph{arXiv preprint arXiv:1603.01547}, 2016.

\bibitem[Ko{\v{c}}isk{\`y} et~al.(2018)Ko{\v{c}}isk{\`y}, Schwarz, Blunsom,
  Dyer, Hermann, Melis, and Grefenstette]{kovcisky2018narrativeqa}
Ko{\v{c}}isk{\`y}, T., Schwarz, J., Blunsom, P., Dyer, C., Hermann, K.~M.,
  Melis, G., and Grefenstette, E.
\newblock The narrativeqa reading comprehension challenge.
\newblock \emph{Transactions of the Association for Computational Linguistics},
  6:\penalty0 317--328, 2018.

\bibitem[Le et~al.(2019{\natexlab{a}})Le, Tran, and Venkatesh]{le2019learning}
Le, H., Tran, T., and Venkatesh, S.
\newblock Learning to remember more with less memorization.
\newblock \emph{arXiv preprint arXiv:1901.01347}, 2019{\natexlab{a}}.

\bibitem[Le et~al.(2019{\natexlab{b}})Le, Tran, and Venkatesh]{le2019neural}
Le, H., Tran, T., and Venkatesh, S.
\newblock Neural stored-program memory.
\newblock \emph{arXiv preprint arXiv:1906.08862}, 2019{\natexlab{b}}.

\bibitem[Le et~al.(2020{\natexlab{a}})Le, Tran, and Venkatesh]{le2020self}
Le, H., Tran, T., and Venkatesh, S.
\newblock Self-attentive associative memory.
\newblock \emph{arXiv preprint arXiv:2002.03519}, 2020{\natexlab{a}}.

\bibitem[Le et~al.(2020{\natexlab{b}})Le, Le, Venkatesh, and
  Tran]{le2020hierarchical}
Le, T.~M., Le, V., Venkatesh, S., and Tran, T.
\newblock Hierarchical conditional relation networks for video question
  answering.
\newblock In \emph{Proceedings of the IEEE Conference on Computer Vision and
  Pattern Recognition}, pp.\  9972--9981, 2020{\natexlab{b}}.

\bibitem[Moscovitch et~al.(2016)Moscovitch, Cabeza, Winocur, and
  Nadel]{moscovitch2016episodic}
Moscovitch, M., Cabeza, R., Winocur, G., and Nadel, L.
\newblock Episodic memory and beyond: the hippocampus and neocortex in
  transformation.
\newblock \emph{Annual review of psychology}, 67:\penalty0 105--134, 2016.

\bibitem[Munkhdalai et~al.(2019)Munkhdalai, Sordoni, Wang, and
  Trischler]{munkhdalai2019metalearned}
Munkhdalai, T., Sordoni, A., Wang, T., and Trischler, A.
\newblock Metalearned neural memory.
\newblock In \emph{Advances in Neural Information Processing Systems}, pp.\
  13331--13342, 2019.

\bibitem[Park et~al.(2020)Park, Choi, and Lee]{park2020distributed}
Park, T., Choi, I., and Lee, M.
\newblock Distributed memory based self-supervised differentiable neural
  computer.
\newblock \emph{arXiv preprint arXiv:2007.10637}, 2020.

\bibitem[Pi et~al.(2019)Pi, Bian, Zhou, Zhu, and Gai]{pi2019practice}
Pi, Q., Bian, W., Zhou, G., Zhu, X., and Gai, K.
\newblock Practice on long sequential user behavior modeling for click-through
  rate prediction.
\newblock In \emph{Proceedings of the 25th ACM SIGKDD International Conference
  on Knowledge Discovery \& Data Mining}, pp.\  2671--2679, 2019.

\bibitem[Rae et~al.(2016)Rae, Hunt, Danihelka, Harley, Senior, Wayne, Graves,
  and Lillicrap]{rae2016scaling}
Rae, J., Hunt, J.~J., Danihelka, I., Harley, T., Senior, A.~W., Wayne, G.,
  Graves, A., and Lillicrap, T.
\newblock Scaling memory-augmented neural networks with sparse reads and
  writes.
\newblock In \emph{Advances in Neural Information Processing Systems}, pp.\
  3621--3629, 2016.

\bibitem[Rae et~al.(2019)Rae, Potapenko, Jayakumar, and
  Lillicrap]{rae2019compressive}
Rae, J.~W., Potapenko, A., Jayakumar, S.~M., and Lillicrap, T.~P.
\newblock Compressive transformers for long-range sequence modelling.
\newblock \emph{arXiv preprint arXiv:1911.05507}, 2019.

\bibitem[Ren et~al.(2019)Ren, Qin, Fang, Zhang, Zheng, Bian, Zhou, Xu, Yu, Zhu,
  et~al.]{ren2019lifelong}
Ren, K., Qin, J., Fang, Y., Zhang, W., Zheng, L., Bian, W., Zhou, G., Xu, J.,
  Yu, Y., Zhu, X., et~al.
\newblock Lifelong sequential modeling with personalized memorization for user
  response prediction.
\newblock In \emph{Proceedings of the International ACM SIGIR Conference on
  Research and Development in Information Retrieval}, pp.\  565--574, 2019.

\bibitem[Santoro et~al.(2018)Santoro, Faulkner, Raposo, Rae, Chrzanowski,
  Weber, Wierstra, Vinyals, Pascanu, and Lillicrap]{santoro2018relational}
Santoro, A., Faulkner, R., Raposo, D., Rae, J., Chrzanowski, M., Weber, T.,
  Wierstra, D., Vinyals, O., Pascanu, R., and Lillicrap, T.
\newblock Relational recurrent neural networks.
\newblock In \emph{Advances in neural information processing systems}, pp.\
  7299--7310, 2018.

\bibitem[Seo et~al.(2016)Seo, Kembhavi, Farhadi, and
  Hajishirzi]{seo2016bidirectional}
Seo, M., Kembhavi, A., Farhadi, A., and Hajishirzi, H.
\newblock Bidirectional attention flow for machine comprehension.
\newblock \emph{arXiv preprint arXiv:1611.01603}, 2016.

\bibitem[Sukhbaatar et~al.(2015)Sukhbaatar, Weston, Fergus,
  et~al.]{sukhbaatar2015end}
Sukhbaatar, S., Weston, J., Fergus, R., et~al.
\newblock End-to-end memory networks.
\newblock In \emph{Advances in Neural Information Processing Systems}, pp.\
  2440--2448, 2015.

\bibitem[Tang \& Wang(2018)Tang and Wang]{tang2018personalized}
Tang, J. and Wang, K.
\newblock Personalized top-n sequential recommendation via convolutional
  sequence embedding.
\newblock In \emph{Proceedings of the Eleventh ACM International Conference on
  Web Search and Data Mining}, pp.\  565--573, 2018.

\bibitem[Vaswani et~al.(2017)Vaswani, Shazeer, Parmar, Uszkoreit, Jones, Gomez,
  Kaiser, and Polosukhin]{vaswani2017attention}
Vaswani, A., Shazeer, N., Parmar, N., Uszkoreit, J., Jones, L., Gomez, A.~N.,
  Kaiser, {\L}., and Polosukhin, I.
\newblock Attention is all you need.
\newblock In \emph{Advances in Neural Information Processing Systems}, pp.\
  5998--6008, 2017.

\bibitem[Weston et~al.(2015)Weston, Bordes, Chopra, Rush, van Merri{\"e}nboer,
  Joulin, and Mikolov]{weston2015towards}
Weston, J., Bordes, A., Chopra, S., Rush, A.~M., van Merri{\"e}nboer, B.,
  Joulin, A., and Mikolov, T.
\newblock Towards ai-complete question answering: A set of prerequisite toy
  tasks.
\newblock \emph{arXiv preprint arXiv:1502.05698}, 2015.

\bibitem[Xiong et~al.(2016)Xiong, Zhong, and Socher]{xiong2016dynamic}
Xiong, C., Zhong, V., and Socher, R.
\newblock Dynamic coattention networks for question answering.
\newblock \emph{arXiv preprint arXiv:1611.01604}, 2016.

\bibitem[Yonelinas(2002)]{yonelinas2002nature}
Yonelinas, A.~P.
\newblock The nature of recollection and familiarity: A review of 30 years of
  research.
\newblock \emph{Journal of memory and language}, 46\penalty0 (3):\penalty0
  441--517, 2002.

\bibitem[Yu et~al.(2019)Yu, Xu, Yu, Yu, Zhao, Zhuang, and
  Tao]{yu2019activitynet}
Yu, Z., Xu, D., Yu, J., Yu, T., Zhao, Z., Zhuang, Y., and Tao, D.
\newblock Activitynet-qa: A dataset for understanding complex web videos via
  question answering.
\newblock In \emph{Proceedings of the American Association for Artificial
  Intelligence}, volume~33, pp.\  9127--9134, 2019.

\bibitem[Zhang et~al.(2021)Zhang, Yao, Zhao, Chua, and Wu]{zhang2021cause}
Zhang, S., Yao, D., Zhao, Z., Chua, T.-S., and Wu, F.
\newblock Causerec: Counterfactual user sequence synthesis for sequential
  recommendation.
\newblock In \emph{Proceedings of the International ACM SIGIR Conference on
  Research and Development in Information Retrieval}, 2021.

\bibitem[Zhang et~al.(2020)Zhang, Zhao, Lin, He,
  et~al.]{zhang2020counterfactual}
Zhang, Z., Zhao, Z., Lin, Z., He, X., et~al.
\newblock Counterfactual contrastive learning for weakly-supervised
  vision-language grounding.
\newblock \emph{Advances in Neural Information Processing Systems},
  33:\penalty0 18123--18134, 2020.

\bibitem[Zhou et~al.(2019)Zhou, Mou, Fan, Pi, Bian, Zhou, Zhu, and
  Gai]{zhou2019deep}
Zhou, G., Mou, N., Fan, Y., Pi, Q., Bian, W., Zhou, C., Zhu, X., and Gai, K.
\newblock Deep interest evolution network for click-through rate prediction.
\newblock In \emph{Proceedings of the American Association for Artificial
  Intelligence}, volume~33, pp.\  5941--5948, 2019.

\end{thebibliography}
\bibliographystyle{icml2021}

\clearpage
\appendix

\section{Task-Specific Reason Models} \label{app.1}
In this section, we introduce the task-specific reason model ${\mathcal R}_{\Omega}({\bf M}, {\bf Q})$, where ${\bf M}$ is the built memory and ${\bf Q}$ is the given query.
Specifically, we first model the query feature ${\bf q} \in \mathbb{R}^{d_{model}}$ by a task-specific encoder. For the synthetic task, the given query ${\bf Q}$ is a one-hot vector and we directly obtain ${\bf q}$ by an embedding layer. For long-sequence text and video QA tasks, the query ${\bf Q}$ is a sentence and we apply a bi-directional GRU to learn the sentence feature ${\bf q}$. As for the recommendation task with long sequences, the given query is a target item with the unique id and we likewise learn an embedding layer to obtain the feature ${\bf q}$.

Next, we develop the multi-hop attention-based reasoning on rehearsal memory ${\bf M}$. 
Concretely, at each step $c$, we capture the importance memory feature ${\bf e}^c \in \mathbb{R}^{d_{x}}$ from ${\bf M}$ based on the current query ${\bf q}^{c-1}$ using an attention method, given by
\begin{eqnarray}
\begin{aligned}
& \gamma^c_{k} = {\bf w}_c^{\top}{\rm tanh}({\bf W}_1^c{\bf q}^{c-1} + {\bf W}_2^c{\bf m}_k + {\bf b}^c), \\  
& {\hat \gamma^c_{k}} = \frac{{\rm exp}(\gamma^c_{k})}{\sum_{j=1}^{K} {\rm exp}(\gamma^c_{j})}, \ {\bf e}^c = \sum_{k=1}^{K} {\hat \gamma^c_{k}} {\bf m}_{k}, \nonumber
\end{aligned}
\end{eqnarray}
where ${\bf W}_1^c \in \mathbb{R}^{d_{model} \times d_{model}}$ , ${\bf W}_2^c  \in \mathbb{R}^{d_{model} \times d_x} $ and ${\bf b}^c  \in \mathbb{R}^{d_{model}} $ are the projection matrices and bias. And ${\bf w}_c^{\top}$ is the row vector.
We then produce the next query ${\bf q}^{c} = {\bf W}^q[{\bf e}^c; {\bf q}^{c-1}] \in \mathbb{R}^{d_{model}}$, where ${\bf W}^q \in \mathbb{R}^{d_{model} \times (d_x+d_{model})}$ is the projection matrix and ${\bf q}^0$ is the original ${\bf q}$. After C steps, we obtain the reason feature ${\bf q}^{C}$. The hyper-parameter $C$ is set to 2, 2, 2 and 1 for synthetic experiments, text QA, video QA and sequence recommendation, respectively. 

After it, we design the final reasoning layer for different tasks. For synthetic experiments and long-sequence video QA with fixed answer sets, we directly apply a classification layer to select the answer and develop the cross-entropy loss ${\mathcal L}_{r}$. But the text QA dataset NarrativeQA provides different candidate answers for each query, we first model each candidate feature ${\bf a}_i$ by another bi-directional GRU and then concatenate ${\bf a}_i$ with ${\bf q}^{C}$ to predict the conference score for each candidate. Finally, we also learn the cross-entropy loss ${\mathcal L}_{r}$ based on answer probabilities. As for the sequence recommendation task, we can directly compute a confidence score based on ${\bf q}^{C}$ by a linear layer and build the binary loss function ${\mathcal L}_{r}$.

\begin{table}[t]
    \centering 
    \begin{threeparttable}
    \caption{Performance Comparisons on Synthetic Data. $R_f$=400, $R_l$=200, $R_q$=40, $R_a$=30, $R_c$=5.}
    \label{table:synthetic}
        \begin{tabular}{c|c|c|c}
            \toprule
           {Method} &{Setting} & Early& Later    \\

            \midrule
            Directly Reason & Directly &13.57&13.41\\
            Multi-Hop Reason & Directly &34.38&34.50\\ 
            \midrule
            \midrule
                DNC &Memory-Based&20.56&26.59\\
                NUTM&Memory-Based&24.31&29.71\\
                {STM} & {Memory-Based} &{23.55}& {29.64}  \\
                {DMSDNC}& {Memory-Based}& {24.92} &{30.74} \\
            \midrule
                RM (w/o. rehearsal) & Memory-Based&25.79&31.38\\
                RM & Memory-Based&{\bf 28.42}&{\bf 31.71}\\ 
            \bottomrule

        \end{tabular}
      \end{threeparttable}
\end{table}

\section{Synthetic Experiment} \label{sec:syn}

{\bf Synthetic Dataset.}
We first introduce the setting of the synthetic task.
Here we abstract the general concepts of reasoning tasks (QA/VQA/Recommendation) to construct the synthetic task.  We define the input sequence as a {\bf Stream} and each item in the sequence as a {\bf Fact}, where the {stream} and {fact} can correspond to the text sequence and word token in text QA. We set the number of fact types to $R_f$, that is, each fact can be denoted by a $R_f$-d one-hot vector and obtain the fact feature by a trainable embedding layer. 
Considering reasoning tasks often need to retrieve vital clues related to the query from the given input and then infer the answer, we define the query-relevant facts in the stream as the {\bf Evidence} and regard the {\bf Evidence}-{\bf Query}-{\bf Answer} triple as the {\bf Logic Chain}. Given a stream and a query, we need to infer the answer if the stream contains the evidence. Specifically, we set the number of query types to $R_q$ and each query can be denoted by a $R_q$-d one-hot vector. For each query, we set the number of answer types to $R_a$. That is, there are totally $R_q * R_a$ query-answer pairs and we need to synthesize $R_q * R_a$ corresponding evidences of each pair. Each evidence is denoted by a sequence of facts $\{{\rm fact}_1, \cdots, {\rm fact}_{R_c}\}$, which continuously appear in the input stream. And $R_c$ is the length of the evidence. During the evidence synthesis, we first define 20 different groups and uniformly split these facts and queries to 20 groups. Next, if a query belongs to group $k$, we randomly sample $R_c$ facts from the group as the evidence, and then assign the evidence to a query-answer pair to generate a fixed logic chain.

Eventually, we synthetic 400 data samples for each logic chain to train the models. Each sample contains the input stream with $R_l$ items, a query and an answer. Concretely, we first sample $R_l$ facts as a sequence and then place the evidence in the sequence, where we guarantee each stream-query pair corresponds to a unique answer.

{\bf Baselines and Model Details.} The {\bf Directly Reason} method first models the input stream by RNN to obtain the stream feature, then concatenates the stream feature with the query feature and predicts the answer by a linear layer. 
The {\bf Multi-Hop Reason} method further applies multiple attention layers after RNN-based stream modeling to capture the query-relevant clues. In the main experiment, we set the dataset hyper-parameters $R_f$, $R_l$, $R_q$, $R_a$ and $R_c$ to 400, 200, 40, 30, and 5, respectively. The facts of the evidence may appear in different stages of the input stream. {\bf Early} means the facts appear in the preceding 50\% of the stream and {\bf Later} means the facts appear in the subsequent 50\%. For our rehearsal memory, we set the $d_x$ and $d_{model}$ to 128. The number K of memory slots and length N of segments are set to 20 and 10, respectively. And we sample all other facts as negative items in ${\mathcal L}_{rec}$.

{\bf Evaluation Results.} Table~\ref{table:synthetic} reports the performance comparison between our method and baselines, where RM is the full model and {RM (w/o. rehearsal)} only employs the task-specific reasoning training. Overall, directly reasoning methods have close early and later performance, but memory-based approaches DNC, NUTM, STM, DMSDNC and  RM (w/o. rehearsal) achieve the terrible early performance due to the gradual forgetting. By the self-supervised rehearsal training, our RM  significantly improves the early accuracy and achieves the best memory-based reasoning performance. This fact suggests our proposed memory rehearsal can alleviate the gradual forgetting of early information and make the memory remember critical information from the input stream. Besides, RM (w/o. rehearsal) outperforms other memory-based methods, which indicates our rehearsal memory machine can better memorize the long-term information even without the rehearsal training. Moreover, we can find the Directly Reason approach achieves the worst performance but the Multi-Hop Reason method has a high accuracy, which demonstrates the performance of directly reasoning methods mainly depends on the complicated interaction between the input contents and queries.

\end{document}